\pdfoutput=1

\documentclass[11pt]{article}

\usepackage[preprint]{acl}

\usepackage{times}
\usepackage{latexsym}

\usepackage[T1]{fontenc}

\usepackage[utf8]{inputenc}

\usepackage{microtype}

\usepackage{inconsolata}
\usepackage{multirow}
\usepackage{booktabs}
\usepackage{linguex}
\usepackage{color, colortbl}
\usepackage{graphicx}
\graphicspath{{/Users/piggy/Desktop/MRP2025} }
\usepackage[export]{adjustbox} 
\usepackage{appendix}

\title{Unpacking Ambiguity: The Interaction of Polysemous Discourse Markers and Non-DM Signals}

\author{
  Jingni Wu \\
  Georgetown University \\
  \texttt{jw2175@georgetown.edu}
  \And
  Amir Zeldes \\
  Georgetown University \\
  \texttt{amir.zeldes@georgetown.edu}
}

\usepackage{booktabs,adjustbox}
\usepackage{array}
\usepackage{caption}
\usepackage{amsmath}

\begin{document}
\maketitle
\begin{abstract}
Discourse markers (DMs) like `but' or `then' are crucial for creating coherence in discourse, yet they are often replaced by or co-occur with non-DMs (`in the morning' can mean the same as `then'), and both can be ambiguous (`since' can refer to time or cause). The interaction mechanism between such signals remains unclear but pivotal for their disambiguation. In this paper we investigate the relationship between DM polysemy and co-occurrence of non-DM signals in English, as well as the influence of genre on these patterns. 
Using the framework of eRST, we propose a graded definition of DM polysemy, and conduct correlation and regression analyses to examine whether polysemous DMs are accompanied by more numerous and diverse non-DM signals. Our findings reveal that while polysemous DMs do co-occur with more diverse non-DMs, the total number of co-occurring signals does not necessarily increase. Moreover, genre plays a significant role in shaping DM-signal interactions.
\end{abstract}

\section{Introduction}
Identifying and understanding discourse relations is fundamental to discourse comprehension. Discourse markers (DM) such as `and', `because', and `however' have been widely recognized as the most typical indicator of coherence relations and are also referred to as discourse connectives or cue phrases \cite{forbes2006computing}. Early research focused on DMs as the sole device indicating  relations, and their presence is often used to distinguish  explicit and implicit relations \cite{webber1998anchoring,robaldo2008penn}. In applied Natural Language Processing (NLP) they also remain the focus of research on automatic detection of discourse relation signaling, as evidenced by the series of DISRPT (Discourse Relation Parsing and Treebanking, see \citealt{braud-etal-2024-disrpt}) shared tasks including DM detection as a track. Since such markers come from a closed list, systems can target only these words or phrases \cite{yu-etal-2019-gumdrop}, then focus on disambiguation, with recent system scores achieving over 93\% F1-scores for English \cite{liu-etal-2023-hits}.

However, more recent studies have shown that DMs account for only a small fraction of discourse relations, which can be signaled by \textit{reference} (e.g.~anaphora to indicate \textsc{elaboration}\footnote{Here and below we will assume discourse relation labels commonly used in Rhetorical Structure Theory \cite{MannThompson1988}. Our definition of what constitutes anaphoric reference aligns with \cite{Zeldes2022}.}), \textit{semantic} (antonymy to indicate \textsc{contrast}), \textit{lexical} (`the next day' can indicate temporal \textsc{sequence} like the DM `then'), \textit{morphological} (past followed by present tense can also indicate \textsc{sequence}) and \textit{graphical} cues (e.g.~a question mark signaling a \textsc{question} relation). In this paper we follow the taxonomy of non-DM signal types proposed by \citet{zeldes-etal-2025-erst}, which distinguishes eight major classes with a total of 45 subtypes, illustrated in Table \ref{tab:sig-types}. Such non-DM signals can be crucial for disambiguating otherwise ambiguous DMs, such as `since', which can signal both \textsc{cause} and temporal \textsc{circumstance} relations. Taken together, DMs and such similar non-DM devices are referred to collectively as discourse relation \textit{signals} \cite{das2018rst,das2018signalling,zeldes-etal-2025-erst}. 

\begin{table*}[h!tb]
\centering
    \resizebox{\textwidth}{!}{%
\begin{tabular}{l|l|l}
\toprule
\textbf{signal type} & \textbf{subtypes} &  \textbf{example}  \\
\midrule
 dm & but, then, on the other hand... & \textit{$[$They wanted to$]$ $[$\textcolor{red}{\textbf{but}} couldn't$]$}$_{<adversative-contrast>}$  \\
\midrule

graphical & colon, dash, semicolon & \textit{$[$Let me tell you a story \textcolor{red}{\textbf{:}}$]$}$_{<organization-preparation>}$  \\

 & layout & \textit{$[$Introduction$]$}$_{<organization-heading>}$  \\
 
 & items in sequence & \textit{1. wash $[$2. cut$]$}$_{<joint-list>}$  \\

 & parentheses, quotation marks & \textit{it rained $[$\textcolor{red}{\textbf{(}}and snowed a bit\textcolor{red}{\textbf{)}}$]$}$_{<elaboration-additional>}$  \\

 & question mark & \textit{$[$Did you\textcolor{red}{\textbf{?}}$]$}$_{<topic-question>}$\textit{ No.}  \\
\midrule

lexical & alternate expression & \textit{He agreed. $[$\textcolor{red}{\textbf{That is}} he said yes$]$}$_{<restatement-repetition>}$  \\

 & indicative word/phrase & \textit{They planned a party! $[$That's \textcolor{red}{\textbf{nice}}/\textcolor{red}{\textbf{Can't wait}}!$]$}$_{<evaluation-comment>}$  \\
 \midrule

morphological & mood & \textit{\textcolor{red}{\textbf{Go}} with them $[$I think you should$]$}$_{<explanation-motivation>}$  \\

& tense & \textit{I \textcolor{red}{\textbf{started}} an hour ago, $[$now I\textcolor{red}{\textbf{'m}} resting$]$}$_{<joint-sequence>}$  \\
\midrule

numerical & same count & \textit{$[$\textcolor{red}{\textbf{Two}} reasons.$]$}$_{<organization-preparation>}$\textit{ First…}  \\
\midrule

reference & 
comparative & \textit{$[$I don’t want \textcolor{red}{\textbf{it}}$]$}$_{<adversative-antithesis>}$\textit{ I want \textcolor{red}{\textbf{another one}}.}  \\

& demonstrative / personal & \textit{They met \textcolor{red}{\textbf{Kim}}. $[$\textcolor{red}{\textbf{This person}} / \textcolor{red}{\textbf{she}} was…$]$}$_{<elaboration-additional>}$  \\

& propositional & \textit{\textcolor{red}{\textbf{They met Kim}}. $[$\textcolor{red}{\textbf{This encouner}} was…$]$}$_{<elaboration-additional>}$  \\
\midrule

semantic & antonymy & \textit{Beer is \textcolor{red}{\textbf{cheap}}, $[$wine is \textcolor{red}{\textbf{expensive}}$]$}$_{<adversative-contrast>}$  \\

& attribution source & \textit{$[$\textcolor{red}{\textbf{Kim}} said$]$}$_{<attribution-positive>}$\textit{ they would}  \\

& lexical chain & \textit{it was \textcolor{red}{\textbf{funny}} $[$so they \textcolor{red}{\textbf{laughed}}$]$}$_{<causal-result>}$  \\

& meronymy & \textit{\textcolor{red}{\textbf{The house}} was big, $[$\textcolor{red}{\textbf{the door}} two meters tall$]$}$_{<elaboration-additional>}$  \\

& negation & \textit{Kim danced, $[$Yun did\textcolor{red}{\textbf{n't}} dance$]$}$_{<adversative-contrast>}$  \\

& repetition/synonymy & \textit{They met \textcolor{red}{\textbf{Dr. Kim}}. $[$\textcolor{red}{\textbf{Dr. Kim}}/\textcolor{red}{\textbf{The surgeon}} was…$]$}$_{<elaboration-additional>}$  \\

\midrule

syntactic & infinitival/relative clause & \textit{a plan $[$\textcolor{red}{\textbf{to}} win$]$}$_{<purpose-attribute>}$  \\

& interrupted matrix clause & \textit{$[$I meant --$]$}$_{<orgnization-phatic>}$\textit{ I mean, }  \\

& modified head & \textit{a \textcolor{red}{\textbf{plan}} $[$to win$]$}$_{<purpose-attribute>}$  \\

& nominal modifier & \textit{articles $[$\textcolor{red}{\textbf{explaining}} chess$]$}$_{<elaboration-attribute>}$  \\

& parallel syntactic construction & \textit{\textcolor{red}{\textbf{it's all}} tasty $[$\textcolor{red}{\textbf{it's all}} pretty$]$}$_{<joint-list>}$  \\

& past/present participial clause & \textit{Kim appeared $[$\textcolor{red}{\textbf{dressed}} in black$]$}$_{<elaboration-attribute>}$  \\

& reported speech & \textit{$[$Kim said$]$}$_{<attribution-positive>}$\textit{ \textcolor{red}{\textbf{that}} they \textcolor{red}{\textbf{would}}}  \\

& subject auxiliary inversion & \textit{I would have $[$\textcolor{red}{\textbf{had}} I known$]$}$_{<contingency-condition>}$  \\
\bottomrule
\end{tabular}
}
\caption{Signal types and subtypes, with examples highlighting in red the signal tokens which indicate the relation of the unit in square brackets.}\label{tab:sig-types}
\vspace{-5pt}
\end{table*}


Despite extensive research on DMs and other signals individually, far less attention has been given to their interaction. Prior studies have examined the distribution of DM-signal co-occurrence and explored potential motivations from corpus-based \cite{das2019multiple, crible2020weak} and experimental perspectives \cite{crible2020role,article}. These studies have revealed that DM-signal co-occurrence is influenced by cognitive constraints and information density, and that several factors, such as the ambiguity of DMs \cite{crible2020weak}, the semantics of discourse relations \cite{das2019multiple,crible2020role}, and genres \cite{crible2020weak}, affect the likelihood of co-occurrence. However, the specific mechanisms governing DM-signal interactions remain unclear. In particular, little is known about which conditions favor such co-occurrences, how different signals contribute to disambiguation and the resulting effect, what happens when conflicting signals appear, and how these patterns vary across discourse relations and genres.  

While previous studies have confirmed that polysemous DMs co-occur with additional signals, there has been little systematic analysis of how different types and combinations of non-DM signals help resolve ambiguity. 
This study seeks to bridge this gap by analyzing the distribution, number, type, and co-occurrence patterns of signals with polysemous DMs across genres. We focus on the following research questions:

\begin{enumerate}
\setlength{\itemsep}{3pt}
    \item Are polysemous DMs accompanied by more numerous or more diverse non-DMs?
    \item What are the typical combination strategies for DM and non-DM signals?
    \item Are strategies and distributions general, or are they genre-specific?
\end{enumerate}

Because of their lower information content, we hypothesize that polysemous DMs will exhibit a stronger connection with non-DM prevalence. We also anticipate that different genres will exhibit distinct preferences for specific types of signals for polysemous DMs when resolving DM ambiguity, in part because they involve different prior likelihoods of certain relations. We therefore expect the relationship between DM polysemy and the number and diversity of co-occurring non-DMs to vary by genre. 

\section{Related Work}
Previous studies have demonstrated that discourse relations are frequently signaled not just by DMs, with over 80\% of signaled relations exhibiting some other textual cues, both with and without the presence of accompanying DMs \cite{taboada2013annotation, das2018rst, das2018signalling}. Moreover, it has been found in many cases that multiple signals indicate discourse relations simultaneously \cite{das2018signalling, webber2019penn}. Among these, the combined use of DMs and non-DM signals is particularly common and serves to signal a wide variety of relations \cite{das2019multiple}. For instance, in the following example from the GUM corpus \cite{zeldes2017gum}, `while' functions as a typical DM for the \textsc{concession} relation, which is further reinforced by a lexical chain connecting existing `studies of the psychology of art' with `no work', creating a contrast between previous work that exists and a gap in academic literature:

\ex. [\textbf{While} \underline{studies of the psychology of art} have focused on ... \underline{no work} has been ...]
[Relation: \textsc{Adversative-concession}; DM: `While'; Signal: semantic (lexical chain)] (File: \textit{GUM\_academic\_art} \label{ex:gum-academic})

Although this pattern is very common in academic writing, little attention has been paid to the ways in which ambiguous DMs such as `while' (which can also mean \textit{during a time that...}) resolve to a unique intepretation thanks to co-occurring signals in this manner, and the joint use of DMs and signals remains a complex question. 

Non-DM signals can 1) overlap with DMs in meaning, potentially leading to redundancy, 2) co-occur with DMs but function independently (potentially signaling multiple distinct relations), and 3) may complement DMs in specific types of relations and environments \cite{hoek2018linguistic}. Recent studies have begun to explore the underlying triggers of the \textit{DM + other signals} phenomenon. Das and Taboada (\citeyear{das2019multiple}) suggested that such combinations may arise from the inherent ambiguity of certain DMs which can signal various relations. For example, the DM \textit{and} can mark additive \textsc{list} and temporal \textsc{sequence} relations, among other options, as illustrated in the following examples from GUM:

\ex. [I came home last night \textbf{and} told you.] [Relation: \textsc{Joint-sequence}] (File: \textit{GUM\_conversation\_grounded})
        
\ex. [... borders of our moral \textbf{and} ethical understanding.] [Relation: \textsc{joint-list}] (File: \textit{GUM\_essay\_ghost})

Building on this, researchers have introduced the concept of \textit{marking strength} or \textit{signaling strength} of DMs, which can be assessed by the number and frequency of discourse relations they can signal \cite{asr2012measuring}. \citet{zeldes2020neural-published} proposed the \textit{delta-softmax} metric, which quantifies prediction accuracy degradation for a trained neural model when a word is removed to estimate its signaling strength for a relation, providing empirical validation of an intuitive graded \textit{signaly-ness} phenomenon. For instance, `but' could be significantly less ambiguous than `and' as a DM, in that removing `but' would make the relation much harder to predict than removing `and'. 

This strength directly influences how DMs interact with non-DM signals: it has been suggested that DMs tend to co-occur more frequently with other signals when indicating a wide range of discourse relations \cite{das2019multiple}. In such cases, non-DM signals can play a disambiguation role, helping to clarify the intended relation \cite{crible2020role}. However, although patterns might be typical of specific genres, for example if formal texts prefer stronger and less ambiguous DMs, the association between DMs and other signals has not been found to vary significantly across genres in previous work \cite{crible2020weak}. 

In addition, combinations of DMs and non-DM signals vary across relation types, but they are not necessarily driven by inherent semantics \cite{das2019multiple}. That is to say, certain relations tend to prefer either DM-only or DM-plus-signal combinations. This is partly influenced by the inherent semantics of the discourse relations themselves (e.g., weakly connected sentences), but also appears to reflect an independent pragmatic strategy for ensuring clarity of the writer’s intention.

While prior research has qualitatively identified some factors influencing the co-occurrences of DM and non-DM signals, a systematic analysis of how specific non-DM signals interact with ambiguous DMs across relation types and genres remains underexplored. In particular, the co-occurrence patterns between ambiguous DMs and accompanying signals have not been quantitatively mapped. Using the largest sample of annotated discourse relation signals to date, this study addresses this gap by investigating 1) the types and frequencies of non-DM signals that co-occur with ambiguous DMs, 2) how these combinations vary across genres, and 3) whether certain signal combinations contribute to disambiguating the intended discourse relation.

\section{Data}
This study uses the Georgetown University Multilayer (GUM) Corpus which consists of 16 spoken and written, informal and formal style English text types \cite{zeldes2017gum} (see corpus details in Appendix \ref{sec:gum-detail}). The corpus originally contained Rhetorical Structure Theory (RST) annotations, which were recently extended based on Enhanced Rhetorical Structure Theory (eRST, \citealt{zeldes-etal-2025-erst}), adding annotated DMs and seven types of non-DM signals based on the taxonomy proposed by \citet{das2018rst}, as well as adding multiple concurrent and tree breaking relations edges to the initial RST trees. With over 250,000 tokens, this is currently the largest dataset annotated for DMs and non-DM signals of discourse relations.

Since this study relies on accurate discourse annotations, we also report how quality was assured. Inter-annotator agreement studies on GUM showed F1 scores of 92.3 for DM identification and 90 for relation association (36 docs, 32K tokens). For non-DM signals, many types were automatically derived from gold syntax and coreference annotations, with others manually corrected or added. On a subset of documents, human–human agreement yielded an F1 of about 0.80.

\section{Polysemy of DMs}
The ambiguous nature of DMs arises from their one-to-many relationship with discourse relations. DMs that can signal multiple relations, such as `and', are often described as \textit{weak signals} \cite{asr2012measuring, das2019multiple, crible2020weak}, as they do not map consistently to a single meaning, in contrast to unambiguous DMs such as `despite', which always marks a \textsc{concession}. 

Going beyond previous categorical approaches to such polysemy, we adopt a graded, quantifiable definition of DM polysemy by calculating the Shannon Entropy \cite{shannon1951prediction} of DM meanings, which measures how evenly a DM is distributed across multiple discourse relations. A high entropy score indicates that a DM appears equally in multiple relations, while a lower score means that a DM is used in only one or very few types of discourse relations, or with a strong predominant sense. We expect a high entropy score here for the most polysemous DMs, for example, a high value for DMs like `and' or even `but', and the lowest value for DMs like `despite'. 

Shannon Entropy is calculated by measuring the probability of the DM appearing in each discourse relation. The polysemy score is computed as follows:

\begin{equation}
  \label{eq:shannon entropy}
  H(X) = - \sum_{i=1}^{n} P(x_i) \log_2 P(x_i)
\end{equation}

$x_i$ is the possible discourse relation signaled by a DM, $n$ is the number of distinct relations signaled by the DM, and $P(x_i)$ is the probability of the DM signaling the relation $x_i$.

\section{DM-Signal Co-occurrences}
\subsection{General Distribution}
\begin{figure*}
    \centering
    \includegraphics[width=0.8\textwidth]{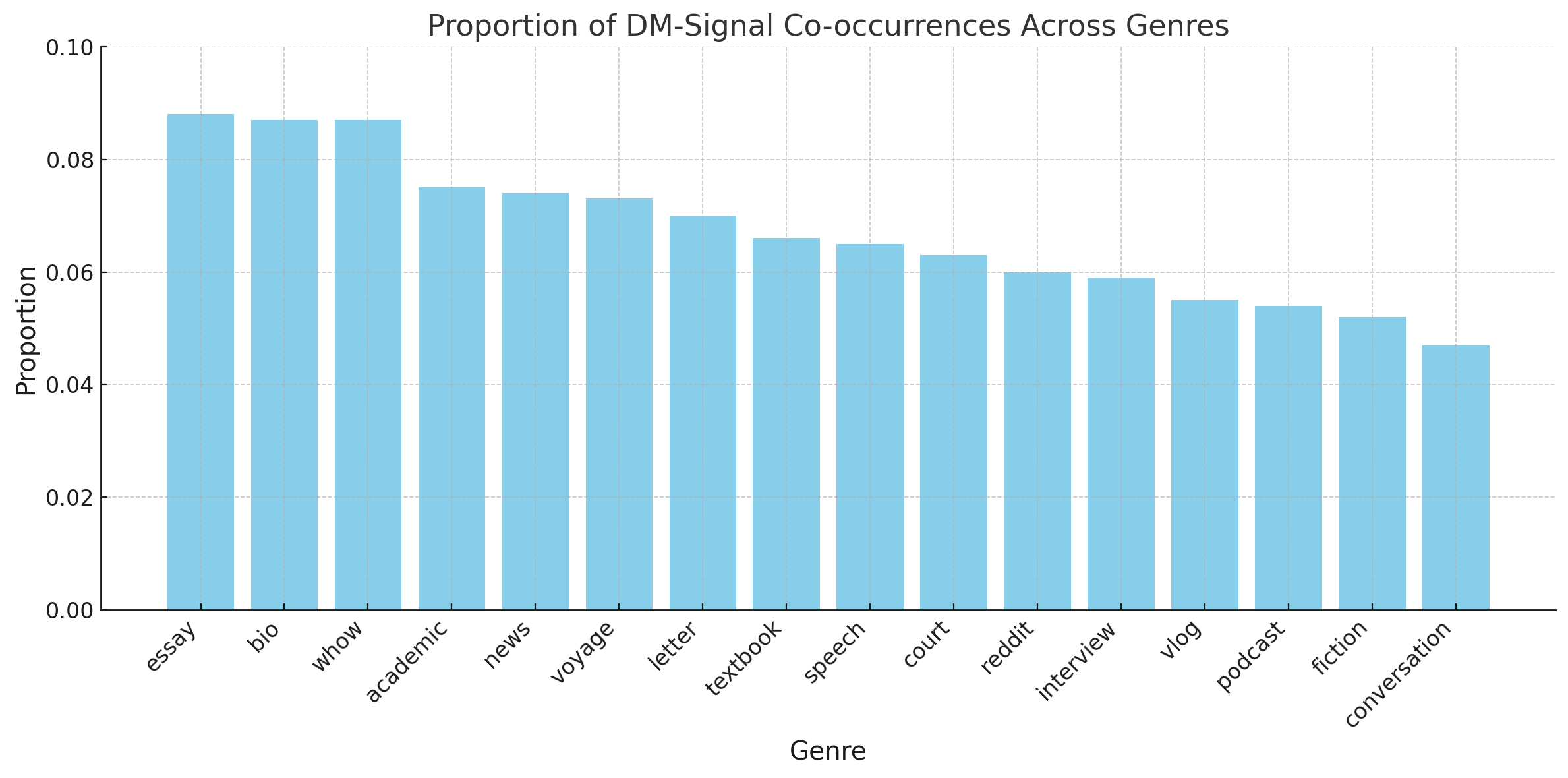}
    \caption{Proportion of DM-Signal co-occurrence across genres}
    \label{fig:dm-signal_co-occurrence}
\end{figure*}

Across 16 genres, 21,435 discourse relations are annotated in our data, of which 1,372 (6.4\%) are indicated by both DMs and non-DM signals. This result aligns fairly closely with Das and Taboada (\citeyear{das2018rst})'s finding for Wall Street Journal news (7.55\%, see also \citealt{LiuZeldes2019}). However as suspected, we observe substantial variation across genres (see Figure \ref{fig:dm-signal_co-occurrence}): \textit{essay} (8.7\%), \textit{bio} (8.6\%), and \textit{whow} (how-to guides form Wikihow, 8.5\%) show a higher proportion of DM-signal co-occurrence, whereas \textit{conversation} has the lowest proportion (3.9\%). 

Among the 1,372 instances of DM-signal co-occurrence, 96\% are marked by DM + 1 signal or DM + 2 signals, while just 3\% are marked by three to four signals. Only a handful of cases include more than five signals (see Table \ref{tab:pattern_dm_signal}). 
\begin{table*}[t] 
    \centering
    \renewcommand{\arraystretch}{1.2} 
    \setlength{\tabcolsep}{4pt} 
    \vspace{10pt} 
    \resizebox{\textwidth}{!}{  
    \begin{tabular}{lc|c|c|c|c|c|c}
        \hline
         & DM + 1 Signal & DM + 2 Signals & DM + 3 Signals & DM + 4 Signals & DM + 5 Signals & DM + 6 Signals & DM + 8 Signals \\
        \hline 
        \hline 
        Total counts & 1092 & 229 & 42 & 6 & 1 & 1 & 1 \\
        \hline 
        Proportion & 79.55\% & 16.7\% & 3.1\% & 0.44\% & 0.07\% & 0.07\% & 0.07\% \\
        \hline 
    \end{tabular}
    } 
    \caption{Pattern of DM + signal combinations in co-occurrences}
    \label{tab:pattern_dm_signal}
    \vspace{10pt} 
\end{table*}

The most commonly used DM in co-occurrence with other signals across genres is the connective `and' (36.6\%), which is generally the most frequently used DM as well. Almost all genres in our corpus employ `and' in DM-signal co-occurrences, except for \textit{academic} where the conjunction `by' is the most common DM favoring non-DM signal accompaniment, as in example \ref{ex:by}, where the DM signaling the \textsc{means} relation is accompanied by the lexical signal `using':

\ex. \textbf{by} \underline{using} a second order Rao and Scott (1981) ... correction\label{ex:by}

\begin{figure*}[t]
    \centering
    \includegraphics[width=\textwidth]{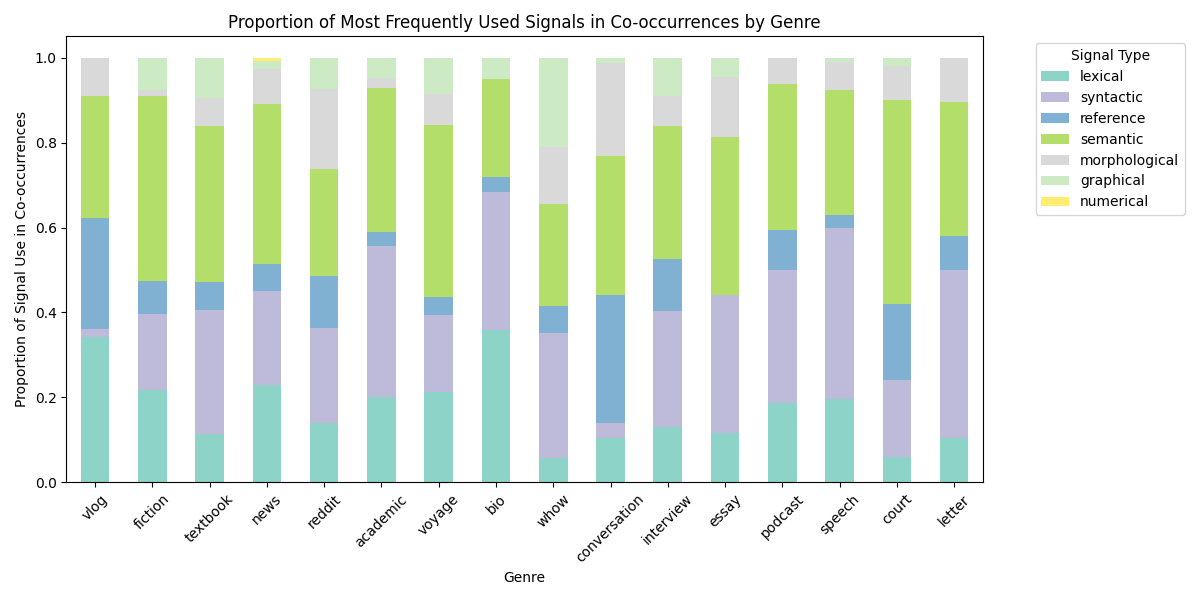} 
    \caption{Proportion of most frequently co-occurring signals by genre}
    \label{fig:proportion of most frequently co-occurred signals per genre}
\end{figure*}

The top three most frequently used signal types in co-occurrences are \textit{semantic}, \textit{syntactic}, and \textit{lexical} across genres, though different genres favor different types of signals, in part due to the format of texts. In spoken genres such as \textit{vlog}, \textit{conversation}, and \textit{court}, there is a large amount of \textit{reference} signals used along with DMs to indicate relations such as \textsc{elaboration} (see Figure \ref{fig:proportion of most frequently co-occurred signals per genre}). Trivially, \textit{graphical} signals such as quotation marks to signal \textsc{attribution} cannot occur in spoken language and are restricted to written data.

\subsection{Polysemous DMs and Signal Patterns}
The DM `so' has the highest polysemy score across all genres in our dataset, while the DM `for' exhibits the most diverse range of accompanying signals (see Table \ref{tab:top5_diverse_dms}). Here, \textit{diversity} \footnote{Since DM frequency varies across genres, we normalized diversity by dividing the number of unique co-occurring signal types by the square root of total DM occurrences. This accounts for diminishing returns and prevents frequent DMs from being unfairly penalized.
} refers to the number of distinct non-DM signal types that co-occur with a given DM, including individual signal types (e.g.~\textit{semantic}) and combinations of multiple types (e.g.~\textit{semantic} + \textit{lexical}). 

\begin{table}[h]  
    \centering
        \begin{tabular}{c|c}
            \toprule
            \textbf{DM} & \textbf{non-DM signal diversity}\\ 
            \midrule
            \textit{for}  & 29.50\\ 
            \textit{and}  & 26.64\\ 
            \textit{if}   & 25.00\\ 
            \textit{by}   & 20.80\\ 
            \textit{when} & 19.00\\ 
            \bottomrule
        \end{tabular}
    \caption{Top 5 DMs with the highest signal diversity}
    \label{tab:top5_diverse_dms}
\end{table}

This raises the question of whether more polysemous DMs tend to co-occur with a greater number of non-DM signals and exhibit more diverse signal patterns, on account of the less consistent mapping of their form to a specific meaning. To answer these questions, we employed fitted regression models to examine the relationship between DM polysemy (independent variable) and two dependent variables: (1) the total number of co-occurring non-DM signals and (2) the diversity of signal types associated with each DM.

Our results, based on both Pearson correlation and regression analyses (see details in Appendix \ref{sec:regression}), suggest that polysemous DMs are more strongly associated with the diversity, rather than the quantity, of accompanying non-DM signals. While we observe a weak but statistically significant correlation between entropy and the total number of co-occurring signals ($r=0.248, p<0.05$), this association does not hold in a multiple regression model where both entropy score and total co-occurring signals are included as predictors of normalized signal diversity. In contrast, entropy remains a significant predictor of normalized diversity, even after controlling for signal quantity ($p<0.001$). This supports the view that more polysemous DMs require more diverse signal patterns rather than just more signals to clarify their discourse functions.

However, the overall explanatory power of entropy score alone is modest (adjusted $R^2 = 0.071$), suggesting that other factors may influence the relationship between DM polysemy and signal diversity. To further explore this, we considered genre as a variable. The regression model (see details in Appendix \ref{sec:regression}) that includes genre and its interaction with entropy score significantly improved model fit ($p < 0.000001$, adjusted $R^2 = 0.090$), suggesting that the effect of DM polysemy on co-occurring signal patterns varies across genres. Notably, genres such as \textit{vlogs} exhibited a significantly stronger positive relationship between DM polysemy and signal diversity, while others like \textit{letter} showed a weaker or even negative trend. This variation highlights that the need for signal diversity in disambiguating polysemous DMs is not uniform, but shaped by genre-specific discourse norms. These genre-specific effects raise the question of what kinds of non-DM signal patterns are employed in each genre, which we address in the next section. 

Looking at patterns rather than counts of signals in more detail, certain signal types consistently co-occur with highly polysemous DMs, suggesting that these signals play a crucial role in disambiguating them. For example, \textit{lexical} and \textit{syntactic} signals frequently appear across multiple cases and are more likely to be combined with other signal types, reinforcing their role in guiding interpretation (see Table \ref{tab:dm_signal_patterns}).
\begin{table*}[h]
    \centering
    \begingroup
    \Large
    \renewcommand{\arraystretch}{2.0}  
    \resizebox{0.9\textwidth}{!}{  
        \begin{tabular}{|c|c|c|}
            \toprule
            \textbf{DM}& \textbf{Top 3 co-occurring Types} & \textbf{Top 3 most frequent combinations} \\ \midrule
            \textit{so}   & morphological, lexical, syntactic  & (lexical + reference), (syntactic + reference + graphical) \\ \hline
            \textit{in}   & syntactic, lexical  & (syntactic + syntactic), (lexical + syntactic + syntactic) \\ \hline
            \textit{with} & semantic, graphical, syntactic  & (reference + semantic), (syntactic + syntactic), (numerical + semantic + semantic) \\ \hline
            \textit{as}   & syntactic, lexical, morphological  & (lexical + semantic) \\ \hline
            \textit{and}  & reference, lexical, semantic  & (reference + graphical), (semantic + semantic), (lexical + syntactic) \\ \bottomrule
        \end{tabular}
    }
    \endgroup
    \caption{Top 5 Polysemous Discourse Markers and Co-occurring Signal Patterns}
    \label{tab:dm_signal_patterns}
\end{table*}

In summary, our hypothesis is partially supported: polysemous DMs are more likely to exhibit diverse combinations of non-DM signal, possibly due to their less stable mapping of form to meaning, but they do not consistently co-occur with a greater number of signals. Given prior evidence that signal co-occurrences vary in quantity across genres (Figure \ref{fig:dm-signal_co-occurrence}), we now turn to investigate the impact of genre variation, and examine the hypotheses within individual genres in the following section.

\subsection{Signal Combinations and Genre Effects}

According to the entropy scores, the most polysemous DMs within each genre are presented in Table \ref{tab:entropy_score_genre}\footnote{When comparing the polysemy across genres, we normalized the entropy score by dividing the raw entropy score by the maximum possible entropy for each DM in each genre.}. Notably, the most ambiguous DMs within each genre differs from those identified as globally most ambiguous. The DM `and' is the most polysemous in six genres, and DM `so' and `as' are the second most ambiguous DMs in eight genres. By contrast, `also' is the most polysemous DM in only one genre. 

\captionsetup[table]{position=bottom} 
\begin{table*}[t]
\centering
\begingroup
  \small
  \setlength{\tabcolsep}{4pt}
  \renewcommand{\arraystretch}{1.1}

  \begin{adjustbox}{max width=\textwidth}
    \begin{tabular}{|l|c|c|c|}
      \hline
      Genre        & DM & Raw entropy & Normalized entropy \\
      \hline\hline
      Court        & and               & 2.85        & 0.61 \\
      Reddit       & so                & 2.59        & 0.60 \\
      Conversation & and               & 2.63        & 0.58 \\
      News         & as                & 2.55        & 0.57 \\
      Fiction      & so                & 2.35        & 0.53 \\
      Voyage       & as                & 2.33        & 0.53 \\
      Interview    & and               & 2.34        & 0.52 \\
      Vlog         & and               & 2.27        & 0.52 \\
      Speech       & so                & 2.20        & 0.51 \\
      Wikihow         & so                & 2.25        & 0.50 \\
      Textbook     & so                & 2.16        & 0.48 \\
      Podcast      & and               & 2.15        & 0.48 \\
      Biography          & also              & 1.90        & 0.44 \\
      Academic     & as                & 1.92        & 0.44 \\
      Letter       & as                & 1.84        & 0.42 \\
      Essay        & and               & 1.64        & 0.40 \\
      \hline
    \end{tabular}
  \end{adjustbox}
\endgroup

\caption{Entropy score of DMs per genre}
\label{tab:entropy_score_genre}
\end{table*}

The non-DM signals that co-occur with polysemous DMs exhibit diverse combination patterns, which vary across genres. A single DM may be more likely to be paired with entirely different signals depending on the genre. For example, the DM `and' is most frequently used with \textit{lexical\_chain} signals (a subtype of \textit{semantic}) signals, see example \ref{ex:lexical_chain}) in nearly all genres, except for \textit{vlog}, \textit{bio}, \textit{whow}, \textit{conversation}, and \textit{podcast} (see Appendix Table~\ref{tab:signal_pattern_and}). Here the lexical relation between the related items `information' and `content' forms a semantic signal next to `and' to indicate that the two clauses are part of a list.

\ex. [The Penn State wiki was never proposed as a source of official \underline{information}, \textbf{and} the university already hosts non-official \underline{content} ...] [Relation: \textsc{joint-list}; DM: `and'; signal: semantic (lexical chain)](File: \textit{GUM\_letter\_wiki)}\label{ex:lexical_chain}

To further assess whether genre systematically affects the distribution of non-DM signals for polysemous DMs, we conducted Chi-Squared Goodness of Fit. For each genre, we compared the signal-type distribution to the global (genre-agnostic) distribution for the same set of DMs. After applying False Discovery Rate (FDR) correction, we found that all 16 genres show statistically significant deviations ($p_\text{corrected} < 0.05$), confirming that genre has a strong effect on the signaling strategies used to support polysemous discourse markers. Genres such as \textit{vlog} and \textit{conversation} exhibited the largest deviations, suggesting that signal use in these genres is especially distinct from the overall norm.

This variation can be attributed to the nature of spoken genres such as \textit{vlogs} and \textit{conversations}, which emphasize audience interaction and shared common ground. In these contexts, indicative words and personal references are more commonly used to enhance engagement and coherence. Similarly, other spoken genres tend to favor \textit{reference} signals, particularly \textit{personal references}, using chains of pronouns to help the audience recall previously mentioned content. Semantic signals in the genre \textit{podcast} show a particularly strong use of \textit{meronymy}, using words in a part-whole relationship alongside the polysemous `and' to indicate elaborations on complex information.

In addition, `and' tends to use combined signals more frequently than other polysemous DMs, dovetailing with our initial hypothesis about non-DMs compensating for ambiguous DMs. Notably, in almost all genres where `and' is the most polysemous DM, it co-occurs with multiple signals, except for the genre \textit{essay}, where it primarily appears by itself or with a single signal type. Among all signal combinations, the most frequent combined signal set for `and' is \textit{reference} + \textit{semantic}, i.e.~anaphora and lexical relations between words in the units joined by `and'. Interestingly, \textit{letter} is the only genre where the most polysemous DM is `as', yet it does not co-occur with any additional non-DM signals. 
Looking at its instances, nearly 65\% are used to indicate \textsc{mode} relations (manner/means), as opposed to only 32.2\% in the rest of the corpus, suggesting that this usage may simply be more predictable as a default in \textit{letters} -- the most common sense in the remaining genres is indicating a temporal \textsc{circumstance}, similarly to `when'. 

\begin{figure*}[t]
    \centering
    \includegraphics[width=\textwidth]{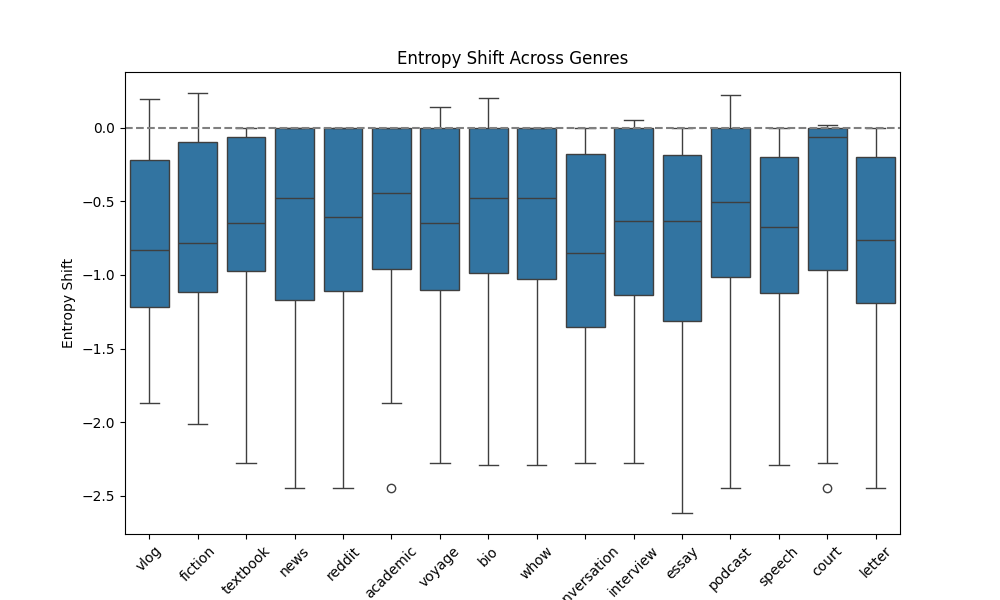} 
    \caption{Entropy shift across genres}
    \label{fig:entropy_shift}
\end{figure*}

Many DMs exhibit reduced polysemy within individual genres compared to their global scores, suggesting that their meaning is more specialized and thus less ambiguous in certain contexts. However, some DMs show substantial variation across genres, potentially requiring a greater variety or higher number of non-DM signals to aid interpretation in specific genres (see Figure~\ref{fig:entropy_shift}). 

To identify DMs whose polysemy varies the most across genres, we compared their normalized within-genre polysemy scores with their global polysemy scores. The top five discourse markers with the largest shifts are `so', `in', `with', `given', and `indeed', which align with the overall polysemy ranking observed earlier. Highly polysemous DMs exhibit greater variance across genres, likely because their multiple meanings make them more adaptable to different discourse needs, which can be disambiguated either by non-DM signals, or simply by their use in a genre with strong priors on expected senses. In contrast to DMs with lower polysemy, which may serve more stable functions, highly polysemous DMs can shift more dramatically depending on genre-specific discourse structures, discourse relation compatibility, communicative conventions, and signaling strategies.

Among the genres, \textit{academic}, \textit{reddit}, and \textit{court} seem to have larger variance, indicating that DMs used in these genres experience the most sizable shifts in polysemy compared to their global usage. These genres may have DMs that behave very differently in terms of polysemy compared to their global usage. In contrast, DMs in \textit{fiction}, \textit{podcast}, and \textit{letter} appear to behave similarly locally and globally. 

In addition, we examined the relationship between the number of co-occurring non-DM signals, the diversity of those signals, and the DM polysemy within each genre. Our global analysis confirms that polysemous DMs tend to co-occur with more diverse signal patterns, however, since the frequency and variety of co-occurring signals differ across genres, we extended this investigation within individual genres to determine whether genre influences this phenomenon. The results indicate that spoken genres such as \textit{court}, \textit{podcast}, and \textit{vlog}, DM polysemy strongly correlates with both the number and diversity of co-occurring non-DM signals, while other genres' results almost align with our findings across genres, that the higher a DM's polysemy score, the more diverse these signal combinations tend to be. This supports our hypothesis that specific genres, particularly spoken contexts and formal or unusual settings (e.g.~courtroom transcripts or academic writing), adopt distinct non-DM signaling strategies which help in the disambiguation of polysemous DMs. 

\section{Conclusion}
This study investigates the relationship between DM polysemy, the number and diversity of co-occurring non-DM signals, and the role of genre in these interactions. Our findings partially support the hypothesis that polysemous DMs exhibit more diverse non-DM signal patterns but do not necessarily co-occur with a greater number of non-DM signals. Moreover, genre greatly shapes DM polysemy, with significant variations in DM entropy and signal usage. Spoken genres (e.g.~\textit{court}, \textit{podcast}, \textit{vlog}) show a stronger dependence on non-DM signals to disambiguate polysemous DMs, while some written genres exhibit little or no correlation. This pattern likely reflects cognitive and interactional pressures in speech, where speakers must maintain fluency under real-time constraints \cite{clark2002speaking} and often deploy additional cues to support coherence. Moreover, DMs in spoken discourse frequently serve interactive functions, such as managing turn-taking or structuring talk \cite{clark2002using}, which may further increase their co-occurrence with diverse signals.

These findings challenge theoretical views of DMs in frameworks that assume that DM marking means we do not need to consider other types of signals, such as in the Penn Discourse Treebank framework, where alternative lexicalizations marking a relation are generally only considered if a DM is absent \cite{prasad-etal-2018-discourse}. They also suggest a consequence for treating DMs and other types of markers as categorically consistent across different types of text: in practice, we find great variation in the extent and types of signaling present based on genre.

On the other hand, some genres exhibit little to no significant correlation among DM polysemy, the number and the diversity of non-DM signals, suggesting that different discourse contexts may impose different constraints on how DMs interact with non-DM signals. Additionally, we identified DMs whose polysemy scores are highly shifted across genres, such as, `so', `in', `with', `given', `indeed', and `while'. This finding suggests that certain polysemous DMs are more sensitive to contextual variation, whereas others maintain stable meanings across different discourse settings. Further research is needed to understand the extent to which the picture of genre variation presented here is comprehensive, which could be carried out with new eRST data on unusual genres that has recently become available (for example in the GENTLE corpus, \citealt{aoyama-etal-2023-gentle}, which includes annotations for poetry, legal writing, and more).

\section{Discussion}
This study does not fully account for the distribution of different discourse relations, which can further shape the observed patterns of polysemy and signal co-occurrence. Prior research has demonstrated that certain non-DMs are more commonly used to disambiguate DMs in specific relations, such as \textit{contrast} and \textit{consequence} \cite{crible2020role}, and different relations may vary in their sensitivity to signals, with some relations being more reliant on co-occurring non-DM cues for disambiguation. Moreover, the compatibility between DMs and specific signals may play a greater role in guiding interpretation than sheer signal frequency. Future work should therefore examine how relation type conditions the use of non-DM signals with polysemous DMs, and expand analysis to larger silver-standard multilayer corpora such as AMALGUM (A Machine Annotated Lookalike of GUM, \citealt{gessler-etal-2020-amalgum}), enriched with automatic annotation of discourse relations and signals, which would be less accurate, but mitigate the problem of data sparseness.


Beyond theoretical implications, these findings also have practical relevance for NLP. Current discourse parsers often treat explicit relations with DMs as straightforward, yet our results show that polysemous markers frequently rely on co-occurring signals. Incorporating such cues could improve discourse relation classification and domain adaptation, while also enhancing explainability in downstream tasks such as summarization or dialogue systems. 

\bibliography{main}

\begin{thebibliography}{28}
\providecommand{\natexlab}[1]{#1}

\bibitem[{Aoyama et~al.(2023)Aoyama, Behzad, Gessler, Levine, Lin, Liu, Peng, Zhu, and Zeldes}]{aoyama-etal-2023-gentle}
Tatsuya Aoyama, Shabnam Behzad, Luke Gessler, Lauren Levine, Jessica Lin, Yang~Janet Liu, Siyao Peng, Yilun Zhu, and Amir Zeldes. 2023.
\newblock \href {https://doi.org/10.18653/v1/2023.law-1.17} {{GENTLE}: A genre-diverse multilayer challenge set for {E}nglish {NLP} and linguistic evaluation}.
\newblock In \emph{Proceedings of the 17th Linguistic Annotation Workshop (LAW-XVII)}, pages 166--178, Toronto, Canada. Association for Computational Linguistics.

\bibitem[{Asr and Demberg(2012)}]{asr2012measuring}
Fatemeh~Torabi Asr and Vera Demberg. 2012.
\newblock Measuring the strength of linguistic cues for discourse relations.
\newblock \emph{Proceedings of the Workshop on Advances in Discourse Analysis and Its Computational Aspects}, pages 33--42.

\bibitem[{Braud et~al.(2024)Braud, Zeldes, Rivi{\`e}re, Liu, Muller, Sileo, and Aoyama}]{braud-etal-2024-disrpt}
Chlo{\'e} Braud, Amir Zeldes, Laura Rivi{\`e}re, Yang~Janet Liu, Philippe Muller, Damien Sileo, and Tatsuya Aoyama. 2024.
\newblock \href {https://aclanthology.org/2024.lrec-main.447/} {{DISRPT}: A multilingual, multi-domain, cross-framework benchmark for discourse processing}.
\newblock In \emph{Proceedings of the 2024 Joint International Conference on Computational Linguistics, Language Resources and Evaluation (LREC-COLING 2024)}, pages 4990--5005, Torino, Italia. ELRA and ICCL.

\bibitem[{Clark(2002)}]{clark2002speaking}
Herbert~H Clark. 2002.
\newblock Speaking in time.
\newblock \emph{Speech Communication}, 36(1-2):5--13.

\bibitem[{Clark and Tree(2002)}]{clark2002using}
Herbert~H Clark and Jean E~Fox Tree. 2002.
\newblock Using uh and um in spontaneous speaking.
\newblock \emph{Cognition}, 84(1):73--111.

\bibitem[{Crible(2020)}]{crible2020weak}
Ludivine Crible. 2020.
\newblock Weak and strong discourse markers in speech, chat, and writing: Do signals compensate for ambiguity in explicit relations?
\newblock \emph{Discourse Processes}, 57(9):793--807.

\bibitem[{Crible and Demberg(2020)}]{crible2020role}
Ludivine Crible and Vera Demberg. 2020.
\newblock The role of non-connective discourse cues and their interaction with connectives.
\newblock \emph{Pragmatics \& Cognition}, 27(2):313--338.

\bibitem[{Das and Taboada(2018{\natexlab{a}})}]{das2018rst}
Debopam Das and Maite Taboada. 2018{\natexlab{a}}.
\newblock {RST Signalling Corpus}: A corpus of signals of coherence relations.
\newblock \emph{Language Resources and Evaluation}, 52:149--184.

\bibitem[{Das and Taboada(2018{\natexlab{b}})}]{das2018signalling}
Debopam Das and Maite Taboada. 2018{\natexlab{b}}.
\newblock Signalling of coherence relations in discourse, beyond discourse markers.
\newblock \emph{Discourse Processes}, 55(8):743--770.

\bibitem[{Das and Taboada(2019)}]{das2019multiple}
Debopam Das and Maite Taboada. 2019.
\newblock Multiple signals of coherence relations.
\newblock \emph{Discours}, (24).

\bibitem[{Forbes-Riley et~al.(2006)Forbes-Riley, Webber, and Joshi}]{forbes2006computing}
Katherine Forbes-Riley, Bonnie Webber, and Aravind Joshi. 2006.
\newblock Computing discourse semantics: The predicate-argument semantics of discourse connectives in {D-LTAG}.
\newblock \emph{Journal of Semantics}, 23(1):55--106.

\bibitem[{Gessler et~al.(2020)Gessler, Peng, Liu, Zhu, Behzad, and Zeldes}]{gessler-etal-2020-amalgum}
Luke Gessler, Siyao Peng, Yang Liu, Yilun Zhu, Shabnam Behzad, and Amir Zeldes. 2020.
\newblock \href {https://aclanthology.org/2020.lrec-1.648/} {{AMALGUM} {--} a free, balanced, multilayer {E}nglish web corpus}.
\newblock In \emph{Proceedings of the Twelfth Language Resources and Evaluation Conference}, pages 5267--5275, Marseille, France. European Language Resources Association.

\bibitem[{Grisot and Blochowiak(2017)}]{article}
Cristina Grisot and Joanna Blochowiak. 2017.
\newblock \href {https://doi.org/10.1075/pc.17009.gri} {Temporal connectives and verbal tenses as processing instructions: Evidence from {F}rench}.
\newblock \emph{Pragmatics \& Cognition}, 24:404--440.

\bibitem[{Hoek et~al.(2018)Hoek, Zufferey, Evers-Vermeul, and Sanders}]{hoek2018linguistic}
Jet Hoek, Sandrine Zufferey, Jacqueline Evers-Vermeul, and Ted~JM Sanders. 2018.
\newblock The linguistic marking of coherence relations: Interactions between connectives and segment-internal elements.
\newblock \emph{Pragmatics \& Cognition}, 25(2):276--309.

\bibitem[{Liu et~al.(2023)Liu, Fan, and Strube}]{liu-etal-2023-hits}
Wei Liu, Yi~Fan, and Michael Strube. 2023.
\newblock \href {https://doi.org/10.18653/v1/2023.disrpt-1.4} {{HITS} at {DISRPT} 2023: Discourse segmentation, connective detection, and relation classification}.
\newblock In \emph{Proceedings of the 3rd Shared Task on Discourse Relation Parsing and Treebanking (DISRPT 2023)}, pages 43--49, Toronto, Canada. The Association for Computational Linguistics.

\bibitem[{Liu and Zeldes(2019)}]{LiuZeldes2019}
Yang Liu and Amir Zeldes. 2019.
\newblock \href {https://doi.org/10.7275/vh3w-4240} {Discourse relations and signaling information: Anchoring discourse signals in {RST-DT}}.
\newblock In \emph{Proceedings of the Society for Computation in Linguistics (SCiL) 2019}, volume~2, pages 314--317.

\bibitem[{Mann and Thompson(1988)}]{MannThompson1988}
William~C. Mann and Sandra~A. Thompson. 1988.
\newblock {Rhetorical Structure Theory}: Toward a functional theory of text organization.
\newblock \emph{Text}, 8(3):243--281.

\bibitem[{Prasad et~al.(2018)Prasad, Webber, and Lee}]{prasad-etal-2018-discourse}
Rashmi Prasad, Bonnie Webber, and Alan Lee. 2018.
\newblock \href {https://aclanthology.org/W18-4710/} {Discourse annotation in the {PDTB}: The next generation}.
\newblock In \emph{Proceedings of the 14th Joint {ACL-ISO} Workshop on Interoperable Semantic Annotation}, pages 87--97, Santa Fe, New Mexico, USA. Association for Computational Linguistics.

\bibitem[{Robaldo et~al.(2008)Robaldo, Miltsakaki, Lee, Prasad, Dinesh, Webber, and Joshi}]{robaldo2008penn}
Alan Robaldo, Eleni Miltsakaki, Alan Lee, Rashmi Prasad, Nikhil Dinesh, Bonnie Webber, and Aravind Joshi. 2008.
\newblock The {Penn Discourse Treebank} 2.0.
\newblock In \emph{Proceedings of the Sixth International Conference on Language Resources and Evaluation (LREC'08)}, Marrakech, Morocco. ELRA.

\bibitem[{Shannon(1951)}]{shannon1951prediction}
Claude~E Shannon. 1951.
\newblock Prediction and entropy of printed {E}nglish.
\newblock \emph{Bell System Technical Journal}, 30(1):50--64.

\bibitem[{Taboada and Das(2013)}]{taboada2013annotation}
Maite Taboada and Debopam Das. 2013.
\newblock Annotation upon annotation: Adding signalling information to a corpus of discourse relations.
\newblock \emph{Dialogue \& Discourse}, 4(2):249--281.

\bibitem[{Webber et~al.(2019)Webber, Prasad, Lee, and Joshi}]{webber2019penn}
Bonnie Webber, Rashmi Prasad, Alan Lee, and Aravind Joshi. 2019.
\newblock The {Penn Discourse Treebank} 3.0 annotation manual.
\newblock \emph{University of Pennsylvania}, 35:108.

\bibitem[{Webber and Joshi(1998)}]{webber1998anchoring}
Bonnie~Lynn Webber and Aravind~K Joshi. 1998.
\newblock Anchoring a lexicalized tree-adjoining grammar for discourse.
\newblock \emph{arXiv preprint cmp-lg/9806017}.

\bibitem[{Yu et~al.(2019)Yu, Zhu, Liu, Liu, Peng, Gong, and Zeldes}]{yu-etal-2019-gumdrop}
Yue Yu, Yilun Zhu, Yang Liu, Yan Liu, Siyao Peng, Mackenzie Gong, and Amir Zeldes. 2019.
\newblock \href {https://doi.org/10.18653/v1/W19-2717} {{G}um{D}rop at the {DISRPT}2019 shared task: A model stacking approach to discourse unit segmentation and connective detection}.
\newblock In \emph{Proceedings of the Workshop on Discourse Relation Parsing and Treebanking 2019}, pages 133--143, Minneapolis, MN. Association for Computational Linguistics.

\bibitem[{Zeldes(2017)}]{zeldes2017gum}
Amir Zeldes. 2017.
\newblock The {GUM} corpus: Creating multilayer resources in the classroom.
\newblock \emph{Language Resources and Evaluation}, 51(3):581--612.

\bibitem[{Zeldes(2022)}]{Zeldes2022}
Amir Zeldes. 2022.
\newblock \href {https://doi.org/10.5210/dad.2022.102} {Can we fix the scope for coreference? problems and solutions for benchmarks beyond {OntoNotes}}.
\newblock \emph{Dialogue \& Discourse}, 13(1):41--62.

\bibitem[{Zeldes et~al.(2025)Zeldes, Aoyama, Liu, Peng, Das, and Gessler}]{zeldes-etal-2025-erst}
Amir Zeldes, Tatsuya Aoyama, Yang~Janet Liu, Siyao Peng, Debopam Das, and Luke Gessler. 2025.
\newblock \href {https://doi.org/10.1162/coli_a_00538} {{eRST}: A signaled graph theory of discourse relations and organization}.
\newblock \emph{Computational Linguistics}, 51(1):23--72.

\bibitem[{Zeldes and Liu(2020)}]{zeldes2020neural-published}
Amir Zeldes and Yang~Janet Liu. 2020.
\newblock \href {https://journals.uic.edu/ojs/index.php/dad/article/view/11372} {A neural approach to discourse relation signal detection}.
\newblock \emph{Dialogue and Discourse}, 11(2):74--99.

\end{thebibliography}

\newpage
\appendix

\section{GUM Information}
\label{sec:gum-detail}

\begin{table}[ht]
  \centering
  \rowcolors{2}{gray!10}{white}
  \begin{tabular}{llrr}
    \toprule
    \textbf{Text type} & \textbf{Source} & \textbf{Docs} & \textbf{Tokens} \\
    \midrule
    Academic writing     & Various        & 18 & 17,169 \\
    Biographies          & Wikipedia      & 20 & 18,213 \\
    CC Vlogs             & YouTube        & 15 & 16,864 \\
    Conversations        & UCSB Corpus   & 15 & 17,932 \\
    Courtroom transcripts& Various        & 9  & 11,148 \\
    Essays               & Various        & 9  & 10,842 \\
    Fiction              & Various        & 19 & 17,511 \\
    Forum                & reddit         & 18 & 16,364 \\
    How–to guides        & wikiHow        & 19 & 17,081 \\
    Interviews           & Wikinews       & 19 & 18,196 \\
    Letters              & Various        & 12 & 9,989  \\
    News stories         & Wikinews       & 24 & 17,186 \\
    Podcasts             & Various        & 10 & 11,986 \\
    Political speeches   & Various        & 15 & 16,720 \\
    Textbooks            & OpenStax       & 15 & 16,693 \\
    Travel guides        & Wikivoyage     & 18 & 16,515 \\
    \midrule
    \textbf{Total GUM}   &                 & 255 & 250,409 \\
    \bottomrule
  \end{tabular}
  \caption{Overview of GUM corpus by text type.}
  \label{tab:gum-text-types}
\end{table}


\section{Signal Patterns for "and" by Genre}
\begin{table*}[h]
    \centering
    \caption{Signal Patterns for "and" by Genre}
    \label{tab:signal_pattern_and}
    \resizebox{\textwidth}{!}{
    \begin{tabular}{l|cc|cc}
        \toprule
        {\textbf{Genre}}& \multicolumn{2}{c|}{Top 3 "and" + 1 signal} & \multicolumn{2}{c}{Top 3 "and" + multiple signals} \\
        & \textbf{Signal Type} & \textbf{Signal Subtype} & \textbf{Signal Type} & \textbf{Signal Subtype} \\
        \midrule
        \multirow{3}{*}{\textbf{vlog}}& lexical & indicative\_word & reference + semantic & personal\_reference + lexical\_chain \\
        & semantic & lexical\_chain & reference + reference & oral\_reference + propositional\_reference \\
        & reference & personal\_reference & reference + semantic & personal\_reference + synonymy \\
        \bottomrule
 \multirow{3}{*}{\textbf{textbook}}& semantic& lexical\_chain& graphical + graphical&items\_in\_sequence + semicolon\\
 & semantic& meronymy& graphical + reference&parentheses + personal\_reference\\
 & graphical& semicolon& semantic + semantic + semantic + semantic&lexical\_chain + lexical\_chain + lexical\_chain + lexical\_chain\\
 \multirow{3}{*}{\textbf{reddit}}& semantic& lexical\_chain & semantic + semantic&lexical\_chain + meronymy\\
 & lexical& indicative\_word& reference + reference + semantic&personal\_reference + propositional\_reference + synonymy\\
 & reference& personal\_reference& reference + semantic + semantic&personal\_reference + lexical\_chain + repetition\\
 \multirow{3}{*}{\textbf{academic}}& semantic& lexical\_chain& reference + semantic&personal\_reference + meronymy\\
 & lexical& indicative\_word& lexical + lexical&indicative\_word + indicative\_word\\
 & semantic& meronymy& graphical + graphical + semantic&items\_in\_sequence + semicolon + meronymy\\
 \multirow{3}{*}{\textbf{voyage}}& semantic& lexical\_chain& semantic + semantic&lexical\_chain + lexical\_chain\\
 & semantic& meronymy& lexical + lexical&indicative\_phrase + indicative\_word\\
 & lexical& indicative\_word& semantic + semantic&lexical\_chain + meronymy\\
 \multirow{3}{*}{\textbf{bio}}& lexical& indicative\_word& lexical + lexical &indicative\_word + indicative\_word\\
 & semantic& lexical\_chain& lexical + lexical &indicative\_phrase + indicative\_word\\
 & lexical& indicative\_phrase& semantic + semantic&lexical\_chain + meronymy\\
 \multirow{3}{*}{\textbf{whow}}& graphical& items\_in\_sequence& semantic + semantic&lexical\_chain + meronymy\\
 & semantic& lexical\_chain& graphical + semantic&items\_in\_sequence + lexical\_chain\\
 & reference& personal\_reference& semantic + semantic&lexical\_chain + lexical\_chain\\
 \multirow{3}{*}{\textbf{conversation}}& reference& personal\_reference& reference + reference&personal\_reference + personal\_reference\\
 & morphological& tense& reference + semantic&personal\_reference + synonymy\\
 & semantic& lexical\_chain& reference + semantic&personal\_reference + lexical\_chain\\
 \multirow{3}{*}{\textbf{fiction}}& semantic& lexical\_chain& semantic + semantic&lexical\_chain + meronymy\\
 & semantic& meronymy& graphical + lexical&semicolon + indicative\_word\\
 & lexical& indicative\_word& semantic + semantic&lexical\_chain + lexical\_chain\\
 \multirow{3}{*}{\textbf{news}}& semantic& lexical\_chain& semantic + semantic&lexical\_chain + meronymy\\
 & semantic& meronymy& lexical + morphological&indicative\_word + tense\\
 & lexical & indicative\_phrase& semantic + semantic&lexical\_chain + lexical\_chain\\
 \multirow{3}{*}{\textbf{interview}}& semantic& lexical\_chain& lexical + lexical +lexical&indicative\_word + indicative\_word + indicative\_word\\
 & reference& personal\_reference& semantic + smeantic + semantic&lexical\_chain + lexical\_chain + meronymy\\
 & graphical& semicolon& semantic + semantic&lexical\_chain + synonymy\\
 \multirow{3}{*}\textbf{essay}& semantic& lexical\_chain& lexical + lexical + lexical &indicative\_word + indicative\_word + indicative\_word\\
 & semantic& meronymy& &\\
 & lexical& alternate\_expression& &\\
 \multirow{3}{*}{\textbf{podcast}}& semantic& meronymy& reference + reference + semantic&demonstrative\_reference + personal\_reference + meronymy\\
 & reference& personal\_reference& semantic + semantic&lexical\_chain + synonymy\\
 & lexical& indicative\_word& lexical + lexical &indicative\_word + indicative\_word\\
 \multirow{3}{*}{\textbf{speech}}& semantic& lexical\_chain& lexical + lexical &indicative\_word + indicative\_word\\
 & syntactic& parallel\_synatactic\_construction& reference + reference&personal\_reference + personal\_reference\\
 & semantic& meronymy& &\\
 \multirow{3}{*}{\textbf{court}}& semantic& lexical\_chain& reference + semantic&demonstrative\_reference + synonymy\\
 & reference& personal\_reference& reference + reference + semantic&personal\_reference + personal\_reference + synonymy\\
 & semantic& negation& reference + semantic&personal\_reference + lexical\_chain\\
 \multirow{3}{*}{\textbf{letter}}& semantic& lexical\_chain& reference + reference&personal\_reference + personal\_reference\\
 & reference& personal\_reference& &\\
 & semantic& meronymy& &\\
    \end{tabular}
    }
\end{table*}

\section{Regression Results}\label{sec:regression}
\appendix

\begin{table*}[h]
\centering
\caption{Pearson Correlation: Entropy Score and Total Co-occurred Signals}
\label{tab:pearson_entropy_signals}
\begin{tabular}{lcc}
\toprule
\textbf{Variable Pair} & \textbf{Correlation (r)} & \textbf{p-value} \\
\midrule
Entropy Score - Total Co-occurred Signals & 0.248 & 0.0137 \\
\bottomrule
\end{tabular}
\end{table*}

\begin{table*}[h]
\centering
\caption{Model 1: Regression of Entropy Score on Normalized Signal Diversity}
\label{tab:normalized_model_1}
\begin{tabular}{lccc}
\toprule
 & \textbf{Coefficient} & \textbf{Std. Error} & \textbf{p-value} \\
\midrule
\textbf{Intercept}       & 0.850               & 0.040            & $<$0.001 \\
\textbf{Entropy Score}   & 0.112               & 0.039            & 0.005 \\
\midrule
$R^2$                    & \multicolumn{3}{c}{0.081} \\
Adjusted $R^2$           & \multicolumn{3}{c}{0.071} \\
F-statistic              & \multicolumn{3}{c}{8.44 \quad ($p = 0.0046$)} \\
\bottomrule
\end{tabular}
\end{table*}

\begin{table*}[h]
\centering
\caption{Model 2: Regression of Entropy Score and Total Signals on Normalized Signal Diversity}
\label{tab:normalized_model_2}
\begin{tabular}{lccc}
\toprule
 & \textbf{Coefficient} & \textbf{Std. Error} & \textbf{p-value} \\
\midrule
\textbf{Intercept}                      & 0.851               & 0.040            & $<$0.001 \\
\textbf{Entropy Score}                  & 0.123               & 0.040            & 0.003 \\
\textbf{Total Co-occurred Signals}     & -0.0005             & 0.0005           & 0.302 \\
\midrule
$R^2$                    & \multicolumn{3}{c}{0.091} \\
Adjusted $R^2$           & \multicolumn{3}{c}{0.072} \\
F-statistic              & \multicolumn{3}{c}{4.76 \quad ($p = 0.0107$)} \\
\bottomrule
\end{tabular}
\end{table*}

\begin{table*}[h]
\centering
\caption{Model 3: Regression of Within-Genre Entropy and Genre Interaction on Normalized Signal Diversity}
\label{tab:model3}
\begin{tabular}{lccc}
\toprule
\textbf{Coefficient} & \textbf{Coef.} & \textbf{Std. Error} & \textbf{p-value} \\
\midrule
\textbf{Intercept} & 0.866 & 0.034 & <0.001 \\
\textbf{Within-Genre Entropy} & 0.055 & 0.033 & 0.098 \\
\midrule
Entropy × Genre[T.vlog] & 0.132 & 0.050 & 0.009 \\
Entropy × Genre[T.letter] & -0.119 & 0.061 & 0.053 \\
Entropy × Genre[T.conversation] & 0.075 & 0.047 & 0.116 \\
Entropy × Genre[T.reddit] & 0.073 & 0.047 & 0.123 \\
Entropy × Genre[T.fiction] & 0.077 & 0.055 & 0.162 \\
Entropy × Genre[T.speech] & -0.066 & 0.049 & 0.178 \\
\midrule
$R^2$ & \multicolumn{3}{c}{0.121} \\
Adjusted $R^2$ & \multicolumn{3}{c}{0.090} \\
F-statistic & \multicolumn{3}{c}{3.97 ($p < 0.000001$)} \\
\bottomrule
\end{tabular}
\end{table*}

\end{document}